\documentclass[runningheads]{llncs}
\usepackage[T1]{fontenc}

\usepackage{multirow}
\usepackage{amsfonts}
\usepackage{dblfloatfix}

\usepackage{graphicx}
\begin{document}
\title{Bi-MCQ: Reformulating Vision–Language Alignment for Negation Understanding}
\titlerunning{Bi-MCQ: Reformulating Vision–Language Alignment for Negation}

\author{Tae Hun Kim\inst{1}\orcidID{0009-0000-8766-4536} \and
Hyun Gyu Lee\inst{2,}\thanks{Corresponding author.}\orcidID{0000-0002-6123-2556}}
\authorrunning{Kim and Lee}

\institute{Department of Electrical and Computer Engineering, Inha University, Republic of  
Korea\\
\email{ka06052@inha.edu}
\and
College of Medicine, Inha University, Republic of Korea \\
\email{hglee@inha.ac.kr}}
\maketitle             
\begin{abstract}
Recent vision--language models (VLMs) achieve strong zero-shot performance via large-scale image--text pretraining and have been widely adopted in medical image analysis. However, existing VLMs remain notably weak at understanding negated clinical statements, largely due to contrastive alignment objectives that treat negation as a minor linguistic variation rather than a meaning-inverting operator. In multi-label settings, prompt-based InfoNCE fine-tuning further reinforces easy-positive image--prompt alignments, limiting effective learning of disease absence. To overcome these limitations, we reformulate vision–language alignment as a conditional semantic comparison problem, which is instantiated through a bi-directional multiple-choice learning framework (Bi-MCQ). By jointly training Image-to-Text and Text-to-Image MCQ tasks with affirmative, negative, and mixed prompts, our method implements fine-tuning as conditional semantic comparison instead of global similarity maximization. We further introduce direction-specific Cross-Attention fusion modules to address asymmetric cues required by bi-directional reasoning and reduce alignment interference. Experiments on ChestXray14, Open-I, CheXpert, and PadChest show that Bi-MCQ improves negation understanding by up to 0.47 AUC over the zero-shot performance of the state-of-the-art CARZero model, while achieving up to a 0.08 absolute gain on positive–negative combined (PNC) evaluation. Additionally, Bi-MCQ reduces the affirmative--negative AUC gap by an average of 0.12 compared to InfoNCE-based fine-tuning, demonstrating that objective reformulation can substantially enhance negation understanding in medical VLMs.
\end{abstract}

\section{Introduction}

Vision–Language Models (VLMs) pretrained on large-scale image–text pairs have demonstrated remarkable performance across a wide range of visual recognition and multimodal reasoning tasks, often rivaling or even surpassing fully supervised approaches without requiring task-specific annotations \cite{clip,align}. By learning a shared embedding space that aligns visual and linguistic representations, these foundation models exhibit strong generalization capability and have become a central paradigm for downstream pattern recognition problems.

Despite their strong performance, VLMs have been reported to struggle with negated statements and explicit descriptions of concept absence \cite{negclip,crepe,vlm_negation,conclip}. Sentences that differ only in negation are often mapped to nearby regions in the joint embedding space, resulting in ambiguous predictions when presence–absence discrimination is required. This limitation is largely attributed to the contrastive learning paradigm, particularly InfoNCE-based objectives that prioritize global image–text similarity over compositional semantics, thereby failing to explicitly model negation as a meaning-altering operator. As a result, affirmative and negated descriptions of the same concept tend to collapse into similar representations.

This issue can be further amplified during downstream fine-tuning. When pretrained VLMs are fine-tuned on label-supervised datasets while retaining the same InfoNCE-based contrastive objective, absence-dominated multi-label settings may reinforce the limitations observed during pretraining. In particular, fine-tuning repeatedly forms easy positive alignments between images and texts describing concept absence, resulting in low-information training signals. As a consequence, the model’s semantic discrimination ability may improve only in a limited manner.

Medical vision–language models provide a representative setting in which this limitation is particularly pronounced. Medical imaging datasets \cite{mimiccxr,chestxray14,openi,chexpert,padchest}, including chest X-rays, are characterized by severe class imbalance and frequent negated statements, where negation serves as a clinically meaningful operator. In this context, medical VLMs often demonstrate strong performance in recognizing disease-related concepts, while exhibiting vulnerability in distinguishing between disease presence and absence \cite{vecl}.

To address these issues, we propose a contrastive-learning-free fine-tuning strategy based on Bi-Directional Multiple Choice Question (Bi-MCQ) learning. The proposed approach jointly optimizes Image-to-Text and Text-to-Image MCQ tasks, encouraging the model to explicitly discriminate among affirmative, negative, and mixed statements. In addition, we introduce a Bi-Directional Cross-Attention-based Image--Text Alignment module that reformulates image--text interaction from global similarity matching into a disease presence–centric semantic comparison problem.

In summary, this work analyzes the structural limitations of InfoNCE-based learning in fine-tuning medical vision–language models for negation understanding, and proposes Bi-MCQ, a Negation-Aware fine-tuning method that serves as an alternative to contrastive learning. These observations suggest that improving negation understanding may require operationalizing the fine-tuning objective itself, rather than further refining contrastive alignment. Moreover, this approach is not limited to vision–language models in the medical domain, but is also significant in that it can be broadly applied to general vision–language models trained on labeling-based datasets without textual descriptions.

\section{Related Works}
\subsection{Chest X-ray-based Medical Vision–Language Models}

Medical vision–language pre-training (VLP) and vision–language models (VLMs) combining Chest X-ray (CXR) images and radiology reports have been widely explored to mitigate the cost of fine-grained annotation. MedCLIP \cite{medclip} extended zero-shot classification by performing contrastive pre-training on unpaired medical images and clinical texts with domain-specific prompt design. CheXzero \cite{chexzero} aligned CXR images and free-text radiology reports within a CLIP framework and formalized a zero-shot diagnostic setting based on similarity comparison between disease-specific positive and negative prompts, without disease labels or domain-specific fine-tuning.

Subsequent studies highlighted the structured, disease-centric nature of radiology reports. MedKLIP \cite{medklip} reconstructed reports into entity- and triplet-level representations to enable clinically meaningful alignment, reducing redundancy and linguistic variability, while KAD \cite{kad} explicitly guided image–text alignment using medical knowledge.

More recent works argued that cosine similarity-based dual encoders are insufficient for modeling clinical relationships. CARZero \cite{carzero} introduced cross-attention-based image–text interaction, and VECL \cite{vecl} further refined alignment by modeling ternary visual entailment relations using positive and negative mentions. Overall, these studies indicate that improving the clinical reliability of medical VLMs requires learning settings that explicitly encode semantic distinctions such as disease presence, absence, and uncertainty.

\subsection{Negation Understanding in Vision–Language Models}

The inability of VLMs to properly handle negation has emerged as a critical limitation affecting linguistic reliability. NegCLIP \cite{negclip} showed that CLIP-style models struggle with compositional semantics and proposed compositional hard negatives, while CREPE \cite{crepe} introduced a benchmark to evaluate compositional reasoning, including negation.

Vision-Language Models Do Not Understand Negation \cite{vlm_negation} demonstrated that state-of-the-art VLMs achieve near-chance performance under negation-inclusive settings and exhibit strong affirmation bias, suggesting a structural rather than data-driven limitation. Although methods such as CoN-CLIP and CC-Neg \cite{conclip} incorporated negation-aware captions into contrastive learning, persistent performance degradation indicates that negation understanding remains unresolved.

Taken together, negation in VLMs is not merely a data scarcity issue but is fundamentally linked to how image–text relationships are defined and compared, motivating the need for learning and evaluation frameworks that more directly encode negation-induced semantic distinctions.

\section{Method}

\begin{figure}
    \centering
    \includegraphics[width=\linewidth]{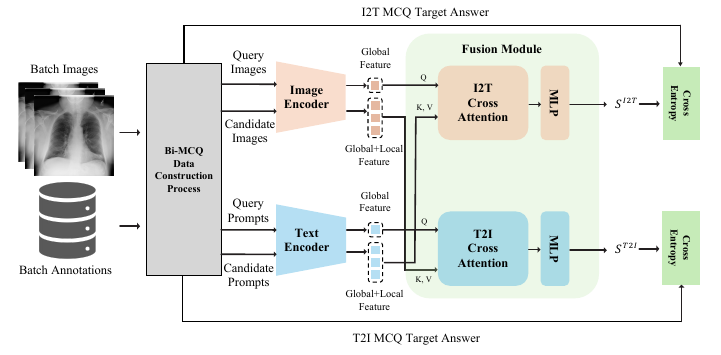}
    \caption{
    Overview of the proposed Bi-MCQ-fine-tuning framework. Annotated chest X-ray data are organized into image-to-text (I2T) and text-to-image (T2I) MCQ batches and processed by image and text encoders. In I2T and T2I settings, modality-specific cross-attention is applied in each direction to obtain similarity scores $S^{\mathrm{I2T}}$ and $S^{\mathrm{T2I}}$, which are optimized using cross-entropy loss.
    }
    \label{fig:placeholder}
\end{figure}

In this section, we present a bi-directional multiple choice(Bi-MCQ) fine-tuning methodology that reformulates annotated chest X-ray datasets for negation-aware adaptation of pretrained medical vision–language models, as illustrated in Fig.~1. We further propose a modality-specific aligned Cross-Attention Fusion Module to support effective bi-directional cross-modal reasoning.

\begin{figure}
    \centering
    \includegraphics[width=\linewidth]{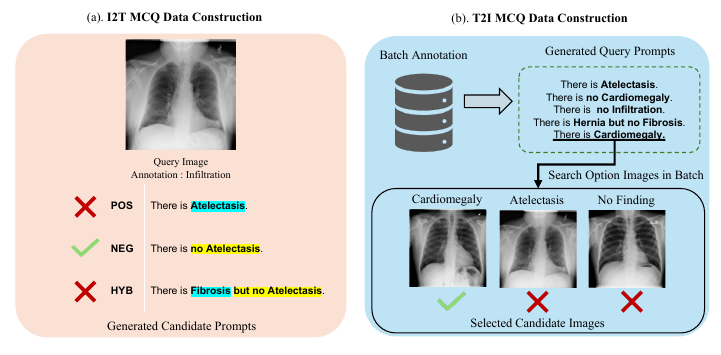}
    \caption{Data construction process for Bi-MCQ-fine-tuning. (a) Image-to-text (I2T) MCQ data construction, where positive, negative, and hybrid candidate prompts are generated for each image in a batch based on its annotation. (b) Text-to-image (T2I) MCQ data construction, where positive, negative, and hybrid query prompts are generated from batch-level annotations and corresponding candidate images are selected from the batch to form MCQ options.}
    \label{fig:placeholder}
\end{figure}

\subsection{Limitations of Contrastive Fine-tuning for Negation Understanding}

When applying medical vision–language models (Medical VLMs) to downstream tasks, a practical approach is to convert labeled disease information into natural language text prompts and use them for fine-tuning. By pairing each image with affirmative and negative prompts describing disease presence or absence, this approach enables prompt-based supervision even when free-form clinical reports are unavailable.

However, learning from negative statements poses a structural challenge. Medical imaging datasets typically contain a high proportion of normal findings or disease-absent samples, resulting in numerous easy-positive alignments between images and generic negation prompts (e.g., “no evidence of disease.”). In such cases, the model repeatedly learns low-information alignment patterns, making it difficult to acquire sufficiently discriminative training signals to finely distinguish disease presence from absence.

This limitation is further amplified by the InfoNCE-based contrastive objective, which maximizes image–text similarity without explicitly modeling the semantic role of negation. Since affirmative and negative prompts share disease terms and differ only by negation cues, contrastive learning tends to emphasize disease-name alignment while treating negation as a minor modifier, thereby hindering meaningful discrimination between disease presence and absence.

\subsection{Data Construction Process for Bi-MCQ-Fine-tuning}

To mitigate the limitations of InfoNCE-based contrastive learning discussed in Section 3.1, we implement labeled medical imaging data into a Multiple Choice Question (MCQ)–based training paradigm. As shown in Fig.~2, our data construction explicitly models bi-directional cross-modal alignment by incorporating both Image-to-Text (I2T) and Text-to-Image (T2I) MCQ settings.

In the I2T MCQ setting, images within a batch are used as queries, and candidate prompts are constructed from the corresponding multi-disease annotations, including affirmative, negative, and mixed statements. Among these candidates, only the prompt that is semantically consistent with the ground-truth annotation is designated as the correct answer, encouraging the model to learn by comparing the semantic differences between sentences containing affirmative and negative expressions.

In the Text-to-Image MCQ (T2I MCQ) setting, all affirmative, negative, and mixed text query prompts that can form a true relationship with images in the batch are generated based on batch-level annotated labels. For each prompt, one image within the batch that is semantically consistent with the disease description is selected as the correct answer, while the remaining images are treated as incorrect options. This setting provides explicit supervision for learning the correspondence between affirmative and negative textual semantics and visual evidence, and prompts for which no appropriate negative images can be identified are excluded from training.

This MCQ-based data construction places affirmative and negative statements in direct competition, structurally removing easy-positive alignments common in prompt-based contrastive learning. As a result, negation is treated as a core semantic factor that must be explicitly resolved to distinguish disease presence from absence.

\subsection{Feature Extraction}

In Bi-MCQ-fine-tuning, both Image-to-Text (I2T) and Text-to-Image (T2I) MCQ settings are jointly considered, with query and candidate embeddings extracted accordingly. Our approach utilizes the image encoder $f_{\mathrm{img}}(\cdot)$ and the text encoder $f_{\mathrm{txt}}(\cdot)$ of a pretrained medical Vision--Language Model, mapping both modalities into a shared $d$-dimensional embedding space.

For the I2T MCQ setting, a query image $I_q$ and its associated candidate text prompts $\{T_i\}_{i=1}^{N}$ are encoded as
\begin{equation}
\mathbf{v}_q = f_{\mathrm{img}}(I_q), \qquad
(\mathbf{t}_i^{\mathrm{g}}, \mathbf{T}_i^{\mathrm{l}}) = f_{\mathrm{txt}}(T_i),
\end{equation}
where $\mathbf{v}_q \in \mathbb{R}^{d}$ denotes the global embedding of the query image, $\mathbf{t}_i^{\mathrm{g}} \in \mathbb{R}^{d}$ represents the sentence-level embedding of the $i$-th candidate prompt, and $\mathbf{T}_i^{\mathrm{l}} \in \mathbb{R}^{L \times d}$ is its token-level embeddings, preserving fine-grained linguistic cues such as negation.

For the T2I MCQ setting, a query text prompt $T_q$ and its associated candidate images $\{I_j\}_{j=1}^{M}$ are encoded as
\begin{equation}
\mathbf{t}_q = f_{\mathrm{txt}}(T_q), \qquad
(\mathbf{v}_j^{\mathrm{g}}, \mathbf{V}_j^{\mathrm{l}}) = f_{\mathrm{img}}(I_j),
\end{equation}
where $\mathbf{t}_q \in \mathbb{R}^{d}$ denotes the global embedding of the query text, $\mathbf{v}_j^{\mathrm{g}} \in \mathbb{R}^{d}$ represents the global embedding of the $j$-th candidate image, and $\mathbf{V}_j^{\mathrm{l}} \in \mathbb{R}^{P \times d}$ denotes its local visual embeddings, which capture fine-grained spatial information for alignment with textual semantics.

\subsection{Image-to-Text Cross-Attention Alignment}

In the Image-to-Text Cross-Attention alignment stage, conditional semantic alignment between a query image embedding and candidate text embeddings is learned. The global image embedding is used as the query, while the global and local text embeddings are used as keys and values in a cross-attention operation:
\begin{equation}
\mathbf{h}_i^{\mathrm{I2T}} =
\mathrm{CrossAttn}_{\mathrm{I2T}}
\Big(
\mathbf{Q}=\mathbf{v}_q,\;
\mathbf{K}=[\mathbf{t}_i^{\mathrm{g}}, \mathbf{T}_i^{\mathrm{l}}],\;
\mathbf{V}=[\mathbf{t}_i^{\mathrm{g}}, \mathbf{T}_i^{\mathrm{l}}]
\Big).
\end{equation}

Through this operation, semantic information provided by each text candidate is selectively aggregated under the image condition, producing representations that reflect how well affirmative, negative, or mixed descriptions are supported by visual evidence. An MLP head is then applied to compute an image--text matching logit:
\begin{equation}
S_i^{\mathrm{I2T}} = \mathrm{MLP}_{\mathrm{I2T}}(\mathbf{h}_i^{\mathrm{I2T}}).
\end{equation}

\subsection{Text-to-Image Cross-Attention Alignment}

In the Text-to-Image Cross-Attention alignment stage, conditional semantic alignment between a query text prompt and candidate image embeddings is learned. The global text embedding is used as the query, while the global and local image embeddings are used as keys and values:
\begin{equation}
\mathbf{h}_j^{\mathrm{T2I}} =
\mathrm{CrossAttn}_{\mathrm{T2I}}
\Big(
\mathbf{Q}=\mathbf{t}_q,\;
\mathbf{K}=[\mathbf{v}_j^{\mathrm{g}}, \mathbf{V}_j^{\mathrm{l}}],\;
\mathbf{V}=[\mathbf{v}_j^{\mathrm{g}}, \mathbf{V}_j^{\mathrm{l}}]
\Big).
\end{equation}

This operation evaluates how the affirmative, negative, or mixed semantics encoded in the text correspond to the visual evidence present in each image. An MLP head is then applied to compute a text--image matching logit:
\begin{equation}
S_j^{\mathrm{T2I}} = \mathrm{MLP}_{\mathrm{T2I}}(\mathbf{h}_j^{\mathrm{T2I}}).
\end{equation}

\subsection{Bi-MCQ-Fine-Tuning Objective}

Based on the matching logits obtained from each direction, Bi-MCQ-fine-tuning formulates both I2T and T2I tasks as multi-class classification problems over candidate sets. For each MCQ, the model is trained to select the candidate that is most semantically compatible with the given query.

For the Image-to-Text MCQ, let the logit vector over text candidates be denoted as
$\mathbf{S}^{\mathrm{I2T}} = \{ S_i^{\mathrm{I2T}} \}_{i=1}^{N}$,
and let $y^{\mathrm{I2T}}$ be the index of the correct text prompt. The loss is defined as
\begin{equation}
\mathcal{L}_{\mathrm{I2T}} =
\mathrm{CrossEntropy}(\mathbf{S}^{\mathrm{I2T}}, y^{\mathrm{I2T}}).
\end{equation}

Similarly, for the Text-to-Image MCQ, given the logit vector
$\mathbf{S}^{\mathrm{T2I}} = \{ S_j^{\mathrm{T2I}} \}_{j=1}^{M}$ and the correct image index $y^{\mathrm{T2I}}$, the loss is defined as
\begin{equation}
\mathcal{L}_{\mathrm{T2I}} =
\mathrm{CrossEntropy}(\mathbf{S}^{\mathrm{T2I}}, y^{\mathrm{T2I}}).
\end{equation}

The final training objective is the sum of the two directional losses:
\begin{equation}
\mathcal{L}_{\mathrm{total}} =
\mathcal{L}_{\mathrm{I2T}} + \mathcal{L}_{\mathrm{T2I}}.
\end{equation}

Importantly, the Image-to-Text and Text-to-Image MCQ losses are backpropagated through separate, direction-specific cross-attention modules, reflecting the distinct conditional reasoning required by each task and preventing representational interference from shared alignment.

Overall, the proposed Bi-MCQ-based fine-tuning strategy explicitly models conditional semantic comparison beyond global similarity alignment. By encouraging direct competition between affirmative, negative, and mixed candidates, it enables more robust learning of fine-grained semantic distinctions and facilitates stable bi-directional semantic alignment centered on disease presence in labeled medical imaging settings.

\section{Experiments}

\subsection{Datasets}

In this study, we evaluate the negation understanding capability and cross-dataset generalization performance of chest X-ray–based medical vision–language models using the ChestXray14 \cite{chestxray14}, Open-I \cite{openi}, CheXpert \cite{chexpert}, and PadChest \cite{padchest} datasets.
ChestXray14 consists of 112,120 chest X-ray images annotated with 14 disease labels. In our experiments, we adopt the official test set released by the NIH, which contains 22,433 images, for performance evaluation. This dataset is the only source used for Bi-MCQ-fine-tuning and also serves as the benchmark for in-domain evaluation. Open-I, CheXpert, and PadChest are employed as external test datasets, each representing different clinical environments and label distributions. For Open-I, evaluation is conducted using the same 14 disease labels defined in ChestXray14. For CheXpert, following standard protocols in prior studies, evaluation is restricted to five major observation categories. For PadChest, we evaluate model performance on five shared disease labels using only the subset of samples with expert-verified manual annotations.

\subsection{Evaluation Metrics}

All experiments are quantitatively evaluated using the Area Under the ROC Curve (AUC), Matthews Correlation Coefficient (MCC), and F1 score for multi-label chest X-ray classification. AUC is a widely adopted metric that provides a robust measure of overall discriminative ability under severe class imbalance, which is common in medical imaging datasets. MCC considers true positives, true negatives, false positives, and false negatives simultaneously, and is particularly suitable for evaluating prediction reliability in datasets dominated by normal samples. The F1 score is additionally reported to assess the balance between precision and recall.

\subsection{Implementation Details}

We compare Bi-MCQ–based fine-tuning with conventional InfoNCE-based fine-tuning strategies on MedKLIP \cite{medklip}, KAD \cite{kad}, and CARZero \cite{carzero}. All models are initialized with publicly available pre-trained weights, with MedKLIP using a ResNet-50 \cite{resnet} image encoder and a ClinicalBERT \cite{clinicalbert} text encoder, KAD using a ResNet-50 image encoder and a PubMedBERT \cite{pubmedbert} text encoder, and CARZero adopting a ViT-B/16 \cite{vit} image encoder with a BioBERT \cite{biobert} text encoder, following their original implementations.

All input images are resized to a resolution of $224 \times 224$. Standard data augmentation techniques, including random horizontal flipping, random affine transformations, and color jittering, are applied during training. Optimization is performed using the Adam optimizer \cite{adam} with a learning rate of $1 \times 10^{-5}$. All experiments are conducted using the PyTorch framework on a single NVIDIA A6000 GPU.

During inference, we employ a positive prompt in the form of There is [disease].'' and a negative prompt in the form of There is no [disease].’’ Through this setup, we evaluate disease classification performance separately for each prompt formulation. In addition, we introduce a Positive–Negative Combined (PNC) inference scheme, which directly compares the outputs of the positive and negative prompts by applying a softmax over their inferred logits, thereby predicting whether an image is more indicative of disease presence or absence.


\begin{table*}[!b]
\renewcommand{\arraystretch}{1.25}
\caption{Cross-dataset evaluation of negation-aware generalization under Bi-MCQ-fine-tuning. Each dataset reports POS / NEG / PNC AUCs. The best result for each metric within the same model is highlighted in bold.}
\label{tab:cross_dataset_pos_neg_pnc}

\makebox[\textwidth][c]{
\resizebox{0.98\textwidth}{!}{
\begin{tabular}{llccc ccc ccc ccc}
\hline
& 
& \multicolumn{3}{c}{\textbf{ChestXray14}} 
& \multicolumn{3}{c}{\textbf{Open-I}} 
& \multicolumn{3}{c}{\textbf{CheXpert}} 
& \multicolumn{3}{c}{\textbf{PadChest}} \\
\cline{3-5}\cline{6-8}\cline{9-11}\cline{12-14}
Model & Setting
& POS & NEG & PNC
& POS & NEG & PNC
& POS & NEG & PNC
& POS & NEG & PNC \\
\hline
CARZero & Zero-shot
& 0.811 & 0.429 & 0.773
& \textbf{0.861} & 0.393 & 0.842
& \textbf{0.923} & 0.491 & \textbf{0.903}
& 0.799 & 0.393 & 0.810 \\

& \textbf{Bi-MCQ-Fine-tuning}
& \textbf{0.851} & \textbf{0.836} & \textbf{0.849}
& 0.858 & \textbf{0.859} & \textbf{0.859}
& 0.896 & \textbf{0.896} & 0.898
& \textbf{0.833} & \textbf{0.811} & \textbf{0.825} \\
\hline
MedKLIP & Zero-shot
& 0.687 & 0.340 & 0.562
& 0.738 & 0.313 & 0.593
& \textbf{0.781} & 0.275 & 0.574
& \textbf{0.515} & 0.480 & \textbf{0.533} \\

& \textbf{Bi-MCQ-Fine-tuning}
& \textbf{0.822} & \textbf{0.822} & \textbf{0.825}
& \textbf{0.776} & \textbf{0.773} & \textbf{0.777}
& 0.770 & \textbf{0.762} & \textbf{0.768}
& 0.484 & \textbf{0.490} & 0.489 \\
\hline
KAD & Zero-shot
& 0.788 & 0.242 & 0.602
& \textbf{0.835} & 0.180 & 0.602
& 0.762 & 0.239 & 0.590
& 0.513 & \textbf{0.470} & \textbf{0.452} \\

& \textbf{Bi-MCQ-Fine-tuning}
& \textbf{0.826} & \textbf{0.822} & \textbf{0.827}
& 0.789 & \textbf{0.776} & \textbf{0.785}
& \textbf{0.771} & \textbf{0.752} & \textbf{0.762}
& \textbf{0.526} & 0.451 & 0.451 \\
\hline
\end{tabular}
}}
\end{table*}

\subsection{Cross-Dataset Generalization of Bi-MCQ-Fine-tuning on ChestXray14}
\label{subsec:cross_dataset_bimcq}

As shown in Table 1, all three models exhibit strong zero-shot performance on positive prompts but near-chance performance on negative prompts, indicating limited sensitivity to disease absence. The inverse POS–NEG relationship observed in MedKLIP and KAD suggests that zero-shot predictions rely largely on surface-level disease name matching rather than semantic reasoning.

Bi-MCQ-fine-tuning consistently improves NEG and PNC performance across ChestXray14, Open-I, and CheXpert, with substantial gains on all metrics in the in-domain ChestXray14 setting. This indicates that Bi-MCQ explicitly enforces disease presence as a discriminative objective, enabling stable reasoning under negated queries and effective transfer to external datasets. In contrast, limited improvements on PadChest highlight the impact of domain mismatch, constraining the effectiveness of ChestXray14-based fine-tuning.

\begin{figure}[!b]
    \centering
    \includegraphics[width=\linewidth]{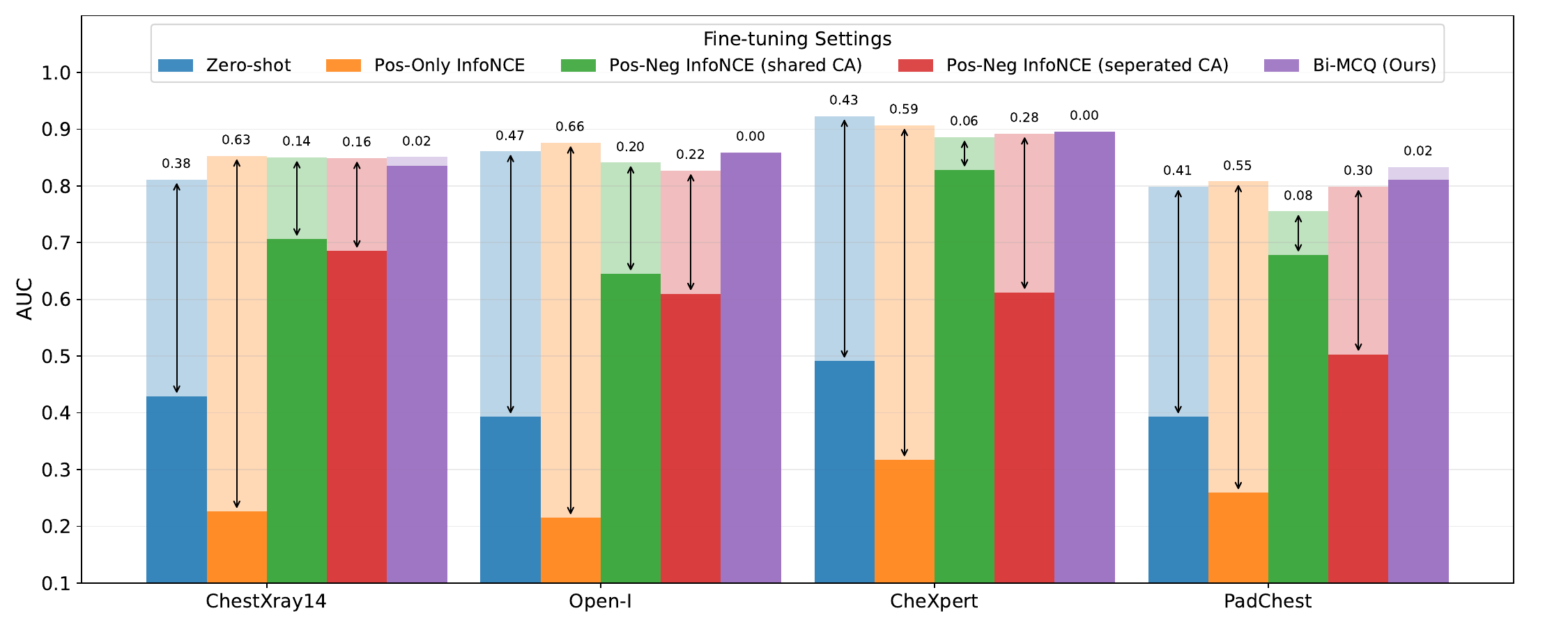}
    \caption{Effect of fine-tuning strategies on CARZero across ChestXray14, Open-I, CheXpert, and PadChest datasets. For each method, the AUC obtained using positive prompts is visualized with higher opacity, while the AUC obtained using negative prompts is shown with lower opacity. The numeric values annotated above the bars indicate the absolute performance gap between positive and negative prompt AUCs, highlighting the robustness of each fine-tuning strategy to prompt polarity.}
    \label{fig:placeholder}
\end{figure}

\subsection{Comparison of Fine-tuning Strategies on CARZero}
\label{subsec:carzero_strategy}

Figure 3 presents a comparison of the average AUC performance of the CARZero model under different fine-tuning strategies, all conducted using an identical ChestXray14 fine-tuning setup. The compared settings include zero-shot classification, InfoNCE loss–based fine-tuning, and Bi-MCQ–based fine-tuning. In addition, under the Pos-Neg setting, we compare a shared cross-attention (shared CA) that jointly models I2T and T2I alignment with a separated cross-attention that explicitly decouples the two alignment directions.

The results reveal that Pos-Only InfoNCE fine-tuning consistently improves performance on positive prompts (POS), while leading to a degradation in negative-prompt (NEG) performance compared to the zero-shot baseline. The Pos-Neg InfoNCE with shared CA setting improves NEG performance across datasets but still exhibits a noticeable performance gap relative to POS. In contrast, Bi-MCQ–based fine-tuning achieves a balanced improvement, maintaining stable and high POS performance while elevating NEG performance to a level comparable with POS.

Meanwhile, applying separated cross-attention to Pos-Neg InfoNCE fine-tuning results in lower performance than Bi-MCQ. This is because, although separated CA structurally decouples Image-to-Text and Text-to-Image alignment, InfoNCE-based training still relies on a global similarity optimization objective that does not explicitly supervise directional or semantic discrimination.
Consequently, the increased structural freedom does not translate into effective modality-specific alignment, resulting in lower negative-prompt performance compared to the shared CA setting, which jointly encodes both alignment directions within a single representation space.

\label{subsec:ablation}
\begin{table*}[!b]
\small{
\renewcommand{\arraystretch}{1.25}
\caption{Ablation study analyzing which components are essential for negation-aware alignment. For each evaluation setting (POS/NEG/PNC), the best value for each metric is highlighted in bold.}
\label{tab:ablation_fusion_freeze}

\makebox[\textwidth][c]{
\resizebox{\textwidth}{!}{
\begin{tabular}{l|l|c|ccc|ccc}
\hline
\textbf{Cross-Attention} & \textbf{Encoder Freeze} & \textbf{Eval} 
& \multicolumn{3}{c}{\textbf{ChestXray14}} 
& \multicolumn{3}{c}{\textbf{PadChest}} \\
\cline{4-6}\cline{7-9}
& & 
& AUC & F1 & MCC 
& AUC & F1 & MCC \\
\hline
Seperated & None & \multirow{4}{*}{POS}
& 0.851 & \textbf{0.362} & \textbf{0.328}
& \textbf{0.833} & 0.178 & 0.181 \\

Seperated & Image &
& \textbf{0.854} & 0.359 & 0.323
& 0.799 & \textbf{0.179} & \textbf{0.184} \\

Seperated & Image + Text &
& 0.849 & 0.358 & 0.324
& 0.809 & 0.177 & 0.183 \\

Shared & None &
& 0.840 & 0.343 & 0.309
& 0.809 & 0.176 & 0.177 \\
\hline
Seperated & None & \multirow{4}{*}{NEG}
& 0.836 & 0.973 & 0.130
& \textbf{0.811} & 0.988 & 0.088 \\

Seperated & Image &
& \textbf{0.839} & 0.973 & \textbf{0.135}
& 0.770 & 0.988 & 0.088 \\

Seperated & Image + Text &
& 0.781 & 0.973 & 0.112
& 0.766 & 0.988 & 0.077 \\

Shared & None &
& 0.838 & 0.973 & 0.118
& 0.792 & 0.988 & \textbf{0.097} \\
\hline
Seperated & None & \multirow{4}{*}{PNC}
& 0.849 & \textbf{0.358} & \textbf{0.324}
& \textbf{0.825} & 0.177 & 0.175 \\

Seperated & Image &
& \textbf{0.853} & 0.354 & 0.321
& 0.794 & \textbf{0.178} & \textbf{0.179} \\

Seperated & Image + Text &
& 0.849 & 0.357 & 0.322
& 0.805 & 0.177 & 0.176 \\

Shared & None &
& 0.846 & 0.344 & 0.310
& 0.805 & \textbf{0.178} & 0.178 \\
\hline
\end{tabular}
}}}
\end{table*}

\begin{table}[!b]
    \centering
    \caption{Ablation study on Bi-MCQ prompt composition with and without mixed prompt augmentation, where all reported values denote AUC.}
    \label{tab:auc_comparison}
    \makebox[\textwidth][c]{
    \resizebox{0.7\textwidth}{!}{
    \begin{tabular}{l|ccc|ccc}
    \hline
    \multirow{2}{*}{\textbf{Setting}} 
    & \multicolumn{3}{c|}{\textbf{ChestXray14}} 
    & \multicolumn{3}{c}{\textbf{Open-I}} \\
    \cline{2-7}
    & POS & NEG & PNC & POS & NEG & PNC \\
    \hline
    Bi-MCQ w.o. mixed 
    & 0.851 & 0.849 & 0.851 
    & 0.845 & 0.846 & 0.847 \\
    Bi-MCQ 
    & 0.851 & 0.836 & 0.849 
    & 0.858 & 0.859 & 0.859 \\
    \hline
    \end{tabular}
    }}
\end{table}

\subsection{Ablation Study}
In Table 2, we present an ablation study to analyze the impact of the Cross-Attention module and encoder-level training on negation understanding performance. Freezing the Image Encoder yields performance comparable to the full training setting on the ChestXray14 dataset, while leading to a decrease in AUC on the PadChest dataset. Furthermore, when both the Image Encoder and Text Encoder are frozen, a substantial drop in AUC is observed even under the NEG setting on ChestXray14, indicating that pretrained encoder representations alone are insufficient to capture the fine-grained semantic distinctions required for negation-aware reasoning.

We further compare different designs of the Cross-Attention Fusion Module to disentangle whether the observed improvement in negation understanding of the proposed fine-tuned model is attributable to the separated CA architecture or to the Bi-MCQ-based fine-tuning approach. 
We intentionally evaluated a shared cross-attention design to test whether the observed gains are attributable to structural decoupling rather than increased representational capacity.
We further compare different designs of the Cross-Attention Fusion Module to disentangle whether the observed improvement in negation understanding originates from the separated alignment structure or from architectural complexity. The results expose that the shared CA structure consistently exhibits lower AUC, F1, and MCC on the ChestXray14 dataset compared to the separated CA structure, and also demonstrates a decrease in AUC on the PadChest dataset. These findings indicate that separating Image-to-Text and Text-to-Image alignment helps mitigate cross-modal interference and enables more precise semantic alignment centered on disease presence or absence.

Table 3 presents a comparison of the average AUC between Bi-MCQ-fine-tuned without mixed prompts and Bi-MCQ incorporating mixed prompts. Since an affirmative-only prompt setting cannot define a correct target prompt for samples without pathological findings, this setting was excluded from the experiment. On the in-domain dataset, ChestXray14, the Bi-MCQ-fine-tuned without mixed prompts, which directly contrasts affirmative and negated prompts, shows a slight improvement in NEG performance. This suggests that learning negation as a direct semantic inversion of affirmative statements is effective in this dataset. In contrast, on the external dataset, Open-I, Bi-MCQ with mixed prompts consistently achieves superior performance across evaluation settings. This indicates that mixed prompts encourage the model to capture finer-grained semantic nuances of the entire textual description, thereby alleviating overfitting. Nevertheless, the performance gap between the two settings remains limited, suggesting that the Bi-MCQ framework itself can achieve stable and competitive performance without a strong reliance on mixed prompt augmentation.

\begin{figure}[t]
\centering
\includegraphics[width=\linewidth]{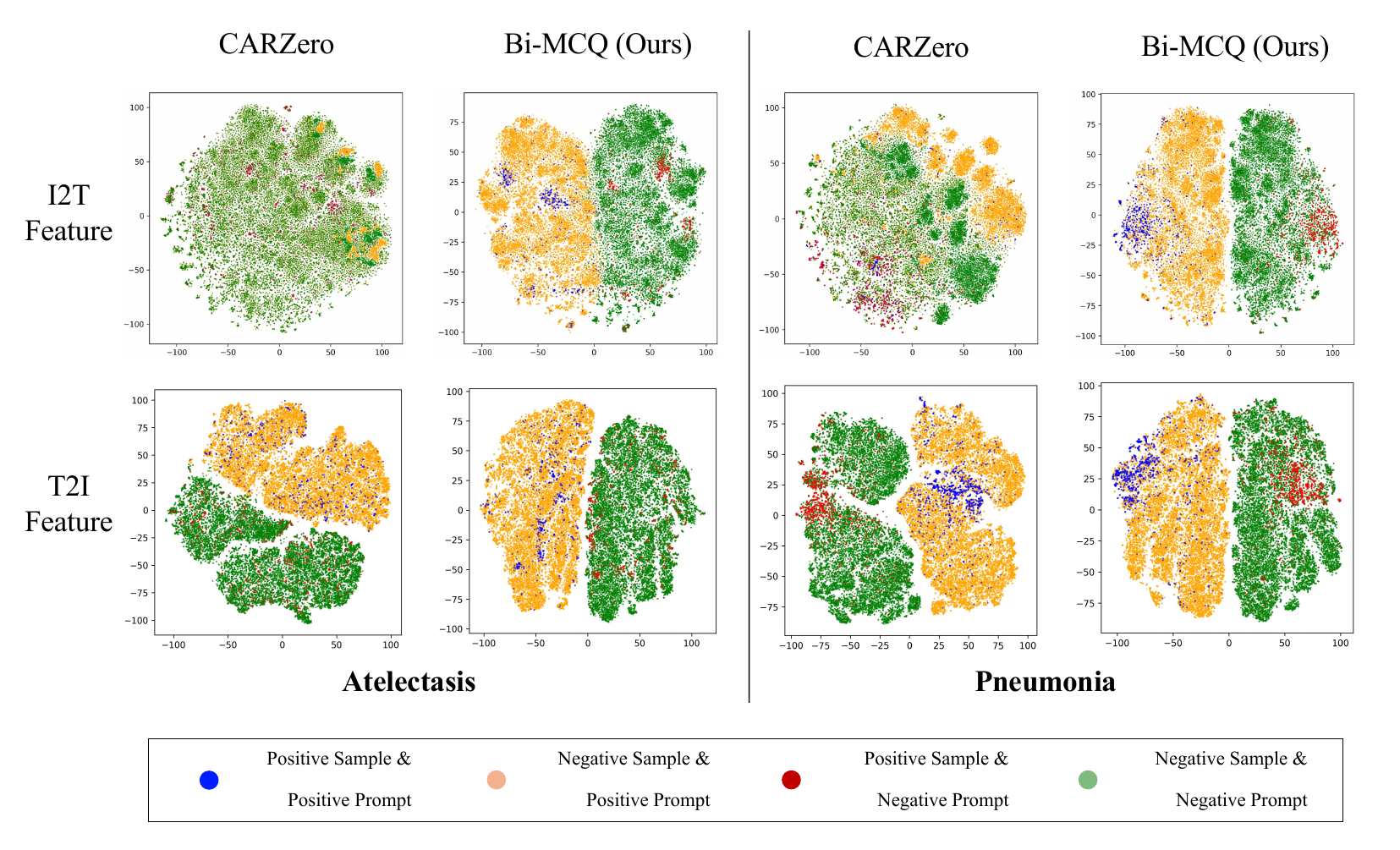}
\caption{t-SNE visualization of feature embeddings of affirmative and negated prompts, highlighting differences in cluster structure associated with prompt polarity.}
\label{fig:tsne}
\end{figure}

\subsection{Visualization}
Fig. 4 provides a qualitative visualization of feature embedding distributions using t-SNE \cite{tsne} for the CARZero model and the Bi-MCQ–fine-tuned model. We emphasize that t-SNE is used solely as a qualitative probe, and that all conclusions in this work are supported by quantitative evaluations. Feature embeddings are extracted from the Image-to-Text (I2T) and Text-to-Image (T2I) cross-attention modules, and are visualized for positive and negative prompts of Atelectasis and Pneumonia, with True and False samples distinguished.

In the I2T feature space, CARZero shows limited separation between positive and negative prompts, with substantial overlap between True and False samples, while the T2I space exhibits relatively clearer separation, reflecting an emphasis on global textual features in its shared alignment module. In contrast, the Bi-MCQ-fine-tuned model consistently separates positive and negative embeddings across both I2T and T2I feature spaces, forming more symmetric embedding structures, suggesting that disease existence is encoded more consistently across alignment directions.

\section{Conclusion}

In this paper, we demonstrate that InfoNCE-based fine-tuning in multi-label chest X-ray settings exhibits a structural limitation in discriminating disease presence from absence. Instead of refining contrastive alignment, this limitation calls for reformulating image–text learning as conditional semantic comparison, which we address by proposing Bi-Directional Multiple Choice Question (Bi-MCQ) based fine-tuning as an alternative to conventional contrastive learning. Furthermore, motivated by the observation that Image-to-Text and Text-to-Image semantic reasoning rely on different cues, we decouple the bi-directional cross-attention alignment modules to mitigate the dilution of subtle semantic differences induced by negated expressions. Experimental results show that Bi-MCQ-based fine-tuning consistently improves negation understanding performance over InfoNCE-based learning state-of-the-art medical vision–language modeling, while maintaining stable generalization performance on external datas-ets. Our Bi-MCQ-based fine-tuning introduces a formal discrepancy between the multi-choice training objective and the inference protocol, where disease predictions are obtained from independently evaluated affirmative and negative prompts. Rather than aiming for exhaustive hypothesis enumeration, Bi-MCQ adopts batch-level candidate sampling, focusing on making negation explicit as a primary semantic dimension. Nevertheless, by explicitly fine-tuning the model under competitive affirmative–negative decision structures, the proposed approach allows inference-time evaluations, including positive–negative combined comparisons, to effectively approximate the learned decision rule, as reflected in consistent performance improvements across datasets. While this study focuses on chest X-rays, by revealing the limitations of existing vision–language model training methodologies, the proposed formulation is not modality-specific and may extend to negation-sensitive vision–language reasoning tasks more broadly.

\begin{credits}
\subsubsection{\ackname} 

\subsubsection{\discintname} 

\end{credits}
%

%
%

\begin{thebibliography}{8}

\bibitem{clip}
Radford, A., Kim, J.W., Hallacy, C., Ramesh, A., Goh, G., Agarwal, S., Sutskever, I.:
Learning transferable visual models from natural language supervision.
In: Proceedings of the 38th International Conference on Machine Learning (ICML),
pp. 8748–8763 (2021)

\bibitem{align}
Jia, C., Yang, Y., Xia, Y., Chen, Y.T., Parekh, Z., Pham, H., Duerig, T.:
Scaling up visual and vision-language representation learning with noisy text supervision.
In: Proceedings of the 38th International Conference on Machine Learning (ICML),
pp. 4904–4916 (2021)

\bibitem{negclip}
Yuksekgonul, M., Bianchi, F., Kalluri, P., Jurafsky, D., Zou, J.:
When and why vision-language models behave like bags-of-words, and what to do about it?
arXiv:2210.01936 (2022)

\bibitem{crepe}
Ma, Z., Hong, J., Gul, M.O., Gandhi, M., Gao, I., Krishna, R.:
CREPE: Can vision-language foundation models reason compositionally?
In: Proceedings of the IEEE/CVF Conference on Computer Vision and Pattern Recognition (CVPR),
pp. 10910–10921 (2023)

\bibitem{vlm_negation}
Alhamoud, K., Alshammari, S., Tian, Y., Li, G., Torr, P.H., Kim, Y., Ghassemi, M.:
Vision-Language Models Do Not Understand Negation.
In: Proceedings of the IEEE/CVF Conference on Computer Vision and Pattern Recognition (CVPR),
pp. 29612–29622 (2025)

\bibitem{conclip}
Singh, J., Shrivastava, I., Vatsa, M., Singh, R., Bharati, A.:
Learning the power of “no”: Foundation models with negations.
In: Proceedings of the IEEE/CVF Winter Conference on Applications of Computer Vision (WACV),
pp. 8002–8012 (2025)

\bibitem{mimiccxr}
Johnson, A.E., Pollard, T.J., Greenbaum, N.R., Lungren, M.P., Deng, C.Y., Peng, Y., Horng, S.:
MIMIC-CXR-JPG, a large publicly available database of labeled chest radiographs.
arXiv:1901.07042 (2019)

\bibitem{chestxray14}
Wang, X., Peng, Y., Lu, L., Lu, Z., Bagheri, M., Summers, R.M.:
ChestX-ray8: Hospital-scale chest X-ray database and benchmarks on weakly-supervised classification and localization of common thorax diseases.
In: Proceedings of the IEEE Conference on Computer Vision and Pattern Recognition (CVPR),
pp. 2097–2106 (2017)

\bibitem{openi}
Demner-Fushman, D., Kohli, M.D., Rosenman, M.B., Shooshan, S.E., Rodriguez, L., Antani, S., McDonald, C.J.:
Preparing a collection of radiology examinations for distribution and retrieval.
Journal of the American Medical Informatics Association 23(2), 304–310 (2016)

\bibitem{chexpert}
Irvin, J., Rajpurkar, P., Ko, M., Yu, Y., Ciurea-Ilcus, S., Chute, C., et al.:
CheXpert: A large chest radiograph dataset with uncertainty labels and expert comparison.
In: Proceedings of the AAAI Conference on Artificial Intelligence,
pp. 590–597 (2019)

\bibitem{padchest}
Bustos, A., Pertusa, A., Salinas, J.M., de la Iglesia-Vayá, M.:
PadChest: A large chest X-ray image dataset with multi-label annotated reports.
Medical Image Analysis 66, 101797 (2020)

\bibitem{vecl}
Wu, W., Yang, J., Zhu, X., Zhang, X., Liu, Z., Li, M., Wu, J.:
Medical contrastive learning of positive and negative mentions.
In: Proceedings of the International Conference on Medical Image Computing and Computer-Assisted Intervention (MICCAI),
pp. 392–401. Springer, Cham (2025)

\bibitem{medclip}
Wang, Z., Wu, Z., Agarwal, D., Sun, J.:
MedCLIP: Contrastive learning from unpaired medical images and text.
In: Proceedings of the Conference on Empirical Methods in Natural Language Processing (EMNLP),
pp. 3876–3887 (2022)

\bibitem{chexzero}
Tiu, E., Talius, E., Patel, P., Langlotz, C.P., Ng, A.Y., Rajpurkar, P.:
Expert-level detection of pathologies from unannotated chest X-ray images via self-supervised learning.
Nature Biomedical Engineering 6(12), 1399–1406 (2022)

\bibitem{medklip}
Wu, C., Zhang, X., Zhang, Y., Wang, Y., Xie, W.:
MedKLIP: Medical knowledge enhanced language-image pre-training for X-ray diagnosis.
In: Proceedings of the IEEE/CVF International Conference on Computer Vision (ICCV),
pp. 21372–21383 (2023)

\bibitem{kad}
Zhang, X., Wu, C., Zhang, Y., Xie, W., Wang, Y.:
Knowledge-enhanced visual-language pre-training on chest radiology images.
Nature Communications 14(1), 4542 (2023)

\bibitem{carzero}
Lai, H., Yao, Q., Jiang, Z., Wang, R., He, Z., Tao, X., Zhou, S.K.:
CARZero: Cross-attention alignment for radiology zero-shot classification.
In: Proceedings of the IEEE/CVF Conference on Computer Vision and Pattern Recognition (CVPR),
pp. 11137–11146 (2024)

\bibitem{resnet}
He, K., Zhang, X., Ren, S., Sun, J.:
Deep residual learning for image recognition.
In: Proceedings of the IEEE Conference on Computer Vision and Pattern Recognition (CVPR),
pp. 770–778 (2016)

\bibitem{clinicalbert}
Alsentzer, E., Murphy, J., Boag, W., Weng, W.H., Jindi, D., Naumann, T., McDermott, M.:
Publicly available clinical BERT embeddings.
In: Proceedings of the 2nd Clinical Natural Language Processing Workshop,
pp. 72–78 (2019)

\bibitem{pubmedbert}
Gu, Y., Tinn, R., Cheng, H., Lucas, M., Usuyama, N., Liu, X., Poon, H.:
Domain-specific language model pretraining for biomedical natural language processing.
ACM Transactions on Computing for Healthcare 3(1), 1–23 (2021)

\bibitem{vit}
Dosovitskiy, A., Beyer, L., Kolesnikov, A., et al.:
An image is worth 16x16 words: Transformers for image recognition at scale.
arXiv:2010.11929 (2020)

\bibitem{biobert}
Lee, J., Yoon, W., Kim, S., Kim, D., Kim, S., So, C.H., Kang, J.:
BioBERT: a pre-trained biomedical language representation model for biomedical text mining.
Bioinformatics 36(4), 1234–1240 (2020)

\bibitem{adam}
Kingma, D.P., Ba, J.:
Adam: A method for stochastic optimization.
arXiv:1412.6980 (2014)

\bibitem{tsne}
van der Maaten, L., Hinton, G.:
Visualizing data using t-SNE.
Journal of Machine Learning Research 9, 2579–2605 (2008)

\end{thebibliography}
%

\end{document}